\documentclass[letterpaper]{article} 
\usepackage{aaai23}  
\usepackage{times}  
\usepackage{helvet}  
\usepackage{courier}  
\usepackage[hyphens]{url}  
\usepackage{graphicx} 
\usepackage{natbib}  
\usepackage{caption} 
\usepackage{algorithm}
\usepackage{algorithmic}

\usepackage{newfloat}
\usepackage{listings}
\DeclareCaptionStyle{ruled}{labelfont=normalfont,labelsep=colon,strut=off} 
\floatstyle{ruled}
\newfloat{listing}{tb}{lst}{}
\floatname{listing}{Listing}
\title{GuidedMixup: An Efficient Mixup Strategy Guided by Saliency Maps}
\author{
    Minsoo Kang\textsuperscript{\rm 1,\rm 2},
    Suhyun Kim\textsuperscript{\rm 2}\thanks{Corresponding author.}
}
\affiliations {
    \textsuperscript{\rm 1} Korea University, Republic of Korea\\
    \textsuperscript{\rm 2} Korea Institute of Science and Technology, Republic of Korea\\
    \{minsoo.kang0918, dr.suhyun.kim\}@gmail.com
}

\usepackage{amsfonts}
\usepackage{verbatim}
\usepackage{times}
\usepackage{amsmath}
\usepackage{multirow}
\usepackage{multicol}
\usepackage{kotex}
\usepackage{amssymb}
\usepackage{pifont}
\usepackage{subfigure}
\usepackage{subfig}
\usepackage{epstopdf}
\usepackage{array}
\usepackage{booktabs}
    
\begin{document}

\maketitle

\begin{abstract}

Data augmentation is now an essential part of the image training process, as it effectively prevents overfitting and makes the model more robust against noisy datasets. Recent mixing augmentation strategies have advanced to generate the mixup mask that can enrich the saliency information, which is a supervisory signal. However, these methods incur a significant computational burden to optimize the mixup mask. From this motivation, we propose a novel saliency-aware mixup method, GuidedMixup, which aims to retain the salient regions in mixup images with low computational overhead. We develop an efficient pairing algorithm that pursues to minimize the conflict of salient regions of paired images and achieve rich saliency in mixup images. Moreover, GuidedMixup controls the mixup ratio for each pixel to better preserve the salient region by interpolating two paired images smoothly. The experiments on several datasets demonstrate that GuidedMixup provides a good trade-off between augmentation overhead and generalization performance on classification datasets. In addition, our method shows good performance in experiments with corrupted or reduced datasets.

\end{abstract}

\section{Introduction}

Improvement and success of deep neural networks have been achieved in various tasks by employing more parameters and larger architectures. However, compared to the capacity increase of deep neural networks, the number of samples in datasets has not kept up. The capacity increase can incur the memorization effect, which appears as an overfitting phenomenon to training data \cite{JMLR:v15:srivastava14a_dropout} and vulnerability to noisy data. 

In order to tackle these issues, many regularization methods have been studied \cite{JMLR:v15:srivastava14a_dropout, huang2016deep}. In particular, data augmentation \cite{krizhevsky2012imagenet, cutout} can effectively regularize the model independently and is now a crucial part of training. The primary purpose of augmentation is to improve the model's generalization ability by generating appropriate unseen data without additional datasets. Recently, a body of research has attempted to develop mixing augmentation techniques. For example, \citet{zhang2018mixup} proposed Mixup, which linearly interpolates images to address the issues of generalization and robustness \cite{stanton2021does}. \citet{yun2019cutmix} proposed CutMix that replaces a random-sized patch of the source image with the same-sized region of the target image.

\begin{figure}[t]
\centering
\includegraphics[width=1.0\linewidth]{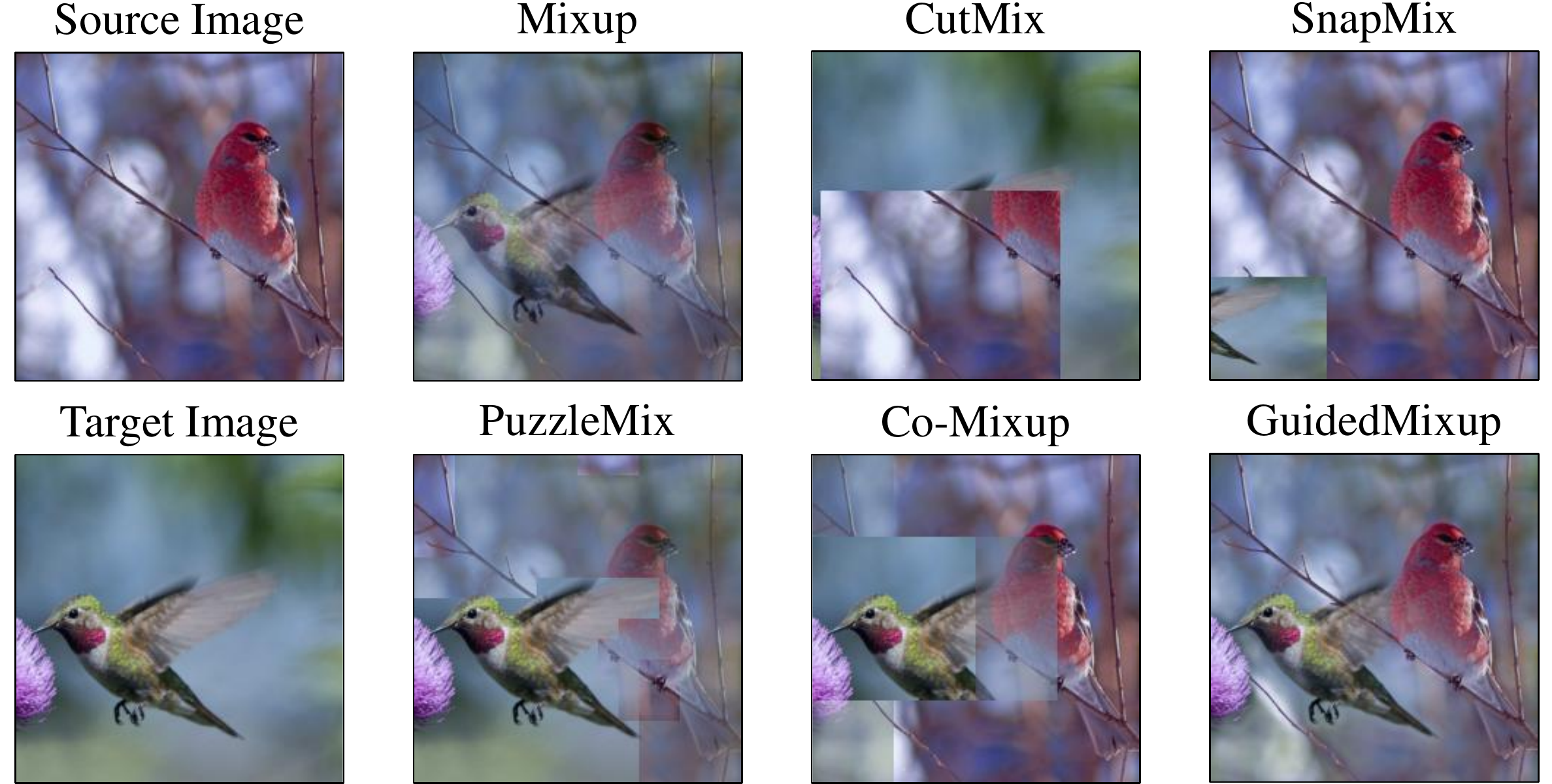}
\caption{The generated images of previous mixup augmentation methods and the proposed GuidedMixup.} \label{examples of mixup}
\end{figure}

These mixup strategies have improved generalization and robustness with lower computational costs. However, as shown in Figure \ref{examples of mixup}, these approaches have limitations in containing enough saliency information (\textit{i.e.}, a supervisory signal). In order to address this issue, there has been another line of studies on properly assigning the saliency information from each image to the mixup images \cite{kim2020puzzle,kim2020co}. PuzzleMix \cite{kim2020puzzle} argued how to generate the mixup mask and optimally transport guided by saliency information. While PuzzleMix rearranges the salient region to the proper location for obtaining maximum saliency, Co-Mixup \cite{kim2020co} finds the optimal pairs and generates the optimal masks simultaneously from the input batch by solving an optimization problem. With the optimal pairs, Co-Mixup was able to obtain richer saliency information in an image than PuzzleMix without the rearrangement. However, saliency-based mixup strategies require a non-trivial computational cost to achieve optimal performance. Furthermore, as the computational inefficiency increases with larger mini-batch, the option of large mini-batch size can be restricted, which may have an optimal point.  

In this regard, we propose GuidedMixup, which achieves both computational efficiency and effective performance by searching the pairs and mixup masks within the input batch. We first formulate a novel pairing optimization problem to minimize the conflict of each salient region in mixup images. Unfortunately, this problem results in a graph-matching problem with higher complexity. To efficiently reduce the complexity, we introduce a greedy pairing algorithm to find the best sets with a large distance between salient regions among the mini-batch images. 

The efficiency of our paring algorithm brings about high flexibility in the choice of the mask unit size. This is in contrast to previous approaches that are limited to using a large block size due to computational cost.
Accordingly, we suggest a mixup mask generation algorithm that maximally leverages saliency information for each pixel and adjusts the mixing ratio pixel by pixel. This can better preserve salient regions of each image within the mixup image and interpolate two images smoothly, which leads to robustness improvement. Figure \ref{examples of mixup} shows the comparison between previous methods and ours.

In the extensive experiments, GuidedMixup non-trivially improves the generalization performance on CIFAR-100 and Tiny-ImageNet. The outperforming results on fine-grained datasets indicate that GuidedMixup can effectively mix images while maintaining the fine details of original images. To validate the efficiency of GuidedMixup, we measure the computational overhead of mixup strategies. We find that GuidedMixup provides good trade-offs between augmentation overhead and generalization performance. Furthermore, since the main purpose of augmentation is to handle overfitting and robustness, we validate our method under the condition of data scarcity and corruption.


\section{Related Work}
Augmentation is one of the regularization methods to address the overfitting problem caused by the gap between the increased capacity and the number of data that is less increased as the model develops. It increases the number of data that gives the diversity of data through various transformations. The most frequently used techniques are horizontal flipping or random cropping \cite{krizhevsky2012imagenet}. Through this, \citet{bishop1995training} has focused on creating a vicinity of a given dataset to improve generalization performance. \citet{bishop2006pattern} also applied a method of giving occlusion to the image by applying various noises or smoothing. In addition to such direct design augmentation, \citet{lemley2017smart} proposed an end-to-end augmentation process, which is Smart Augmentation, that learns an augmentation process consisting of a serial combination of multiple affine transforms. 
\citet{hendrycks2020augmix} proposed AugMix, which interpolates multiple warping augmentation techniques to improve the robustness of the neural network and prevent overfitting.

Compared to data-warping augmentation, the other line of augmentation, which is the mixup-based method, has recently been proposed. \citet{zhang2018mixup} proposed the method named ‘MixUp,’ which creates linearly interpolated virtual training examples with a random mixing ratio. Mixup leads the model to have smoother decision boundaries and ameliorates the robustness against noise data \cite{stanton2021does}. \citet{verma2019manifold} proposed an extended version of Mixup named Manifold Mixup, which interpolates data at the feature level. Also, \citet{yun2019cutmix} suggested CutMix, which replaces a randomly generated spatial patch of the input with another and assigns an area ratio label that is irrelevant to the actual information in the mixed data. SaliencyMix \cite{uddin2020saliencymix} proposed a CutMix-based method so that a random-sized patch is cut from the most salient point in an image.

The randomness in existing methods may generate improper samples, resulting in misleading models. For this reason, generating images appropriate for the model has become mainstream research. SnapMix \cite{huang2021snapmix} proposed the idea of asymmetrically replacing the random size patch of an image with another by utilizing a class activation map (CAM) \cite{zhou2016learning} as the guidance of a semantic label. PuzzleMix \cite{kim2020puzzle} also proposes a mixup method that transports rich salient information to another image by solving the optimization problem. \citet{kim2020co} proposed Co-MixUp, which finds the proper combination of salient regions and pairs simultaneously from the optimal solution of optimization problems.

``Saliency object detection" was first suggested by \citet{730558}, who integrated two separate stages that are 1) distinguishing the most salient object from the background (detecting) and 2) segmenting the specific region of the object (segmenting). We roughly categorize saliency object detection methods in two ways, whether utilizing the neural network or not. \citet{achanta2009frequency} proposed the detection methods that use frequency domain without the neural network.
\citet{hou2007saliency} proposed a spectral residual approach that utilizes conventional signal processing theory to distinguish the objects by approximating background statistics. On the contrary, \citet{simonyan2013deep} proposed the method that generates a saliency map by using the computed gradients from the pre-trained model. \citet{zhao2015saliency} proposed the method to precisely generate the saliency map by simultaneously combining global and local contexts from the pre-trained model.

\begin{figure*}[t]
    \begin{center}
    \includegraphics[width=0.9\textwidth]{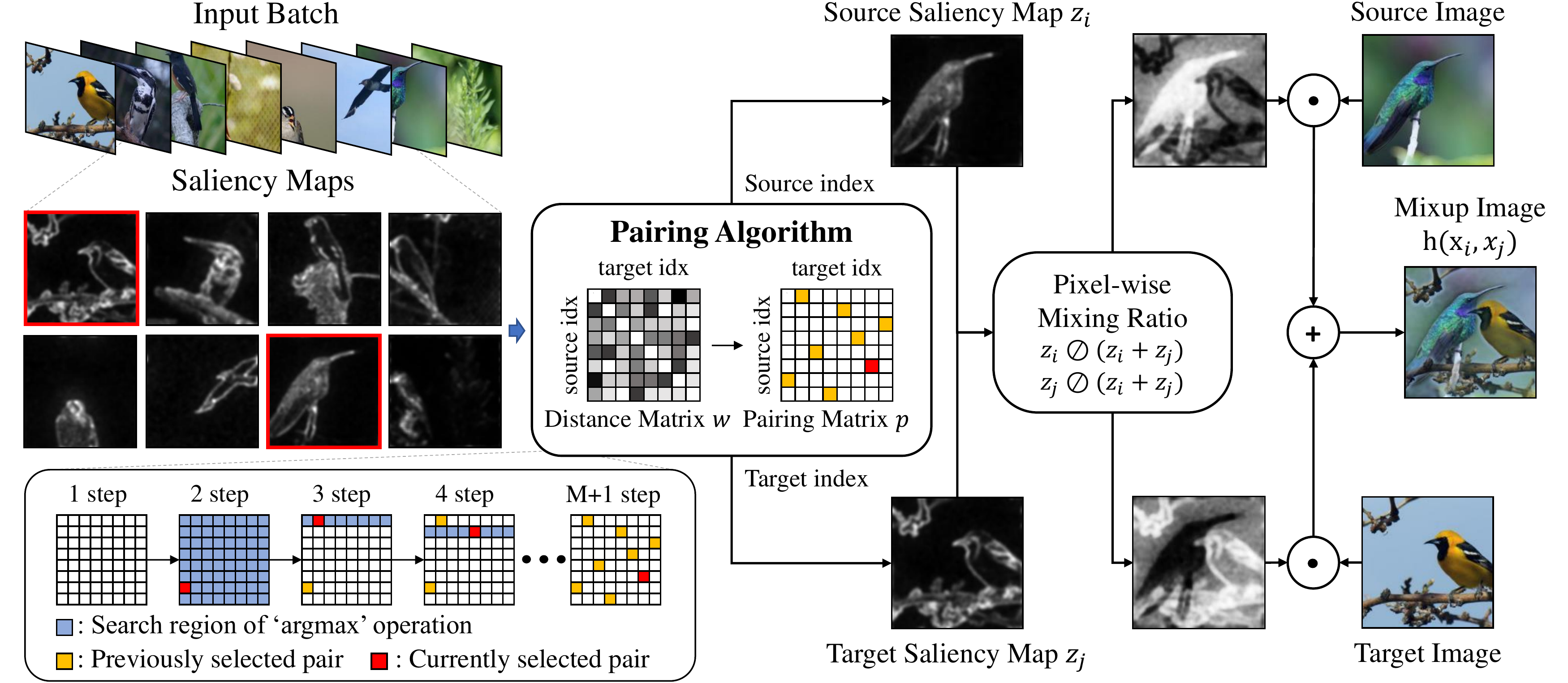}
    \caption{An overall diagram of GuidedMixup. GuidedMixup augmentation pipeline consists of the following 4 steps: 1) Generate saliency maps from mini-batch images. 2) Construct a pairing matrix $p$ from the distance matrix $w$ by utilizing pairing algorithm. 3) To obtain mixup masks, adjust the pixel-wise mixing ratio between each pixel in the source and target saliency map. 4) Final mixup images are obtained as the pixel-wise sum of masked images.}
    \label{method_images}
    \end{center}
\end{figure*}

\section{Preliminary}
Mixing data augmentation methods generate synthetic data $\hat{x}$ and the corresponding label $\hat{y}$ from a given mixup function $\hat{x}=h(x_{s},x_{t})$, where $x_{s},x_{t} \in\mathbb{R}^{N}$ are the source and target image, respectively. 

For example, in the case of input mixup, $h(x_{s},x_{t})=\lambda x_{s}+(1-\lambda)x_{t}$. CutMix uses $h(x_{s},x_{t})= z \odot x_{s}+(1-z)\odot x_{t}$, where $z \in {\{0,1\}}^{N}$ is a binary rectangular-shaped mask which has the same shape of target image $x_{t}\in N$ and $\odot$ represents the point-wise multiplication. SaliencyMix uses the same function as CutMix but obtains the location of the rectangular-shaped mask from the maximum point of a saliency map. PuzzleMix uses $h(x_{s},x_{t})=z\odot\Pi^{\top}_{s}x_{s}+(1-z)\odot\Pi^{\top}_{t}x_{t}$ where $\Pi$ is a transport mask and $z$ is a discrete saliency-based mask. 

For mini-batch size input, a mixup function can be described as the following: $\hat{x}_{B}=H(x_B)$. Co-Mixup uses $H(x_{B}) = \{g(z_{m}\odot x_{B})|m \in B \}$, where $x_{B}$ is a set of images which have mini-batch size $M$ and $z_{m}\in \mathbb{R}^{M\times N}$ is the mask that makes a pair and augments simultaneously. $g$ is a function $\mathbb{R}^{M\times N} \rightarrow \mathbb{R}^{N}$ that conducts the column-wise summation.  
\section{Method}

The central idea of GuidedMixup is to preserve the object's shape as much as possible by maintaining the saliency regions in the mixed images through the pairing method and pixel-wise mixing ratio. A recent study \cite{kim2020co} tries to solve the optimization of mix-matching the inputs to have the maximized saliency measure. As a result, time complexity becomes a burden, which is considering pairs and block-based mask generation simultaneously. 

In order to reduce this overhead, we split the augmentation process into two, pairing algorithm and mixup mask generation, which are both guided by saliency information. In solving an optimization problem, this division reduces the number of inputs that increases complexity, such as the batch size of images and the size of the saliency map. The overall diagram is illustrated in Figure \ref{method_images}. In the following sections, we provide a brief background and introduce our GuidedMixup method in detail.

\subsection{Pairing Algorithm} \label{pairing_algorithm}
The pairing algorithm in GuidedMixup pursues to let the mixup images embrace the salient regions unperturbed as much as possible. The main objective of the pairing algorithm is to choose source and target pairs towards maximizing the saliency measure by the minimum overlap of salient regions from the mini-batch.

To obtain pairs guided by saliency maps, we conduct the Gaussian-blur on the saliency map $s(x_{k})$ to cover the object well. After Gaussian-blurring the saliency map, we conduct the sum-to-1 normalization $z_{k,i}=s(x_{k,i})/\sum_{n=1}^{N}{s(x_{k,n})}$ to prevent the masks from being biased to the specific images having large salient regions. This can lead to fairly defining the relationship between inputs for matching pairs. As a metric of pair compatibility, we utilize the $\ell_{2}$ distance between saliency maps $z_{k}$. Maximizing the distance between the selected pairs corresponds to minimizing the overlap of salient regions and rich supervisory signals in mixup images. To achieve this, we first calculate a matrix $w\in \mathbb{R}^{M\times M}$ that contains the $\ell_{2}$ distance of each saliency map pair. 
And then, we find a pairing matrix $p \in \{0,1\}^{M\times M}$ that maximizes the summation of the element-wise multiplication between the distance matrix $w$ and the pairing matrix $p$. 
The problem can be formulated as follows:
\begin{equation} \label{optimization objective}
    \begin{aligned}
        {\text{maximize}}\quad & {\sum_{(i,j)\in B \times B}{w_{i,j}p_{i,j}}}, \\
        \text{subject to}\quad & {\sum_{i\in B}{p_{i,j}}}=1,\\ 
        & {\sum_{j\in B}{p_{i,j}}}=1,\\
        & p_{i,j}+p_{j,i}\leq1 \quad i, j\in B, \\
        & p_{i,j}=0 \quad \text{if } i=j,\quad i, j\in B.
    \end{aligned}
\end{equation}
All the constraints are diversity terms. The first and second constraints ensure that an image is selected only once as a source or target. The third constraint prevents a symmetric pair $(j, i)$ from being selected when a pair $(i, j)$ is already selected. The fourth constraint avoids an image being paired with itself. 

This optimization problem reduces to a Maximum weight $k$-cycle cover in a complete graph, where $k$ is the number of cycles linked to at least three vertices. We can optimally solve this problem in polynomial time by a maximum matching algorithm \cite{Edmonds1965MaximumMA} through Tutte's reduction \cite{Tutte1954ASP}. However, the algorithm requires high complexity O($M^{3}$) and can be solved only when the edge weight is an integer to conduct graph reduction. Moreover, since the value of $\ell_{2}$ distance is a real number, mapping to a proper integer can be costly. 

As an efficient way to find a satisfactory solution, we propose a greedy algorithm to solve Equation \ref{optimization objective}, described in Algorithm \ref{greedy_pairing_algorithm}. The algorithm draws one cycle cover in a graph by greedily proceeding to the best remaining neighbor at each vertex. It first selects the largest distance from the distance matrix $w$, then selects the largest from the row corresponding to the target index used as the source in turn while satisfying the constraints. The time complexity is O($M^{2}$), which comes from both finding the first maximum and the loop with the `argmax' operation.

Since the algorithm pairs minimally conflicting images, the proposed pairing algorithm can be applied to other mixup methods. In Section \ref{impact_pairing}, we discuss the impact of our pairing algorithm on other mixup baselines.

\begin{algorithm}[t]
    \caption{Greedy Pairing Algorithm}
    \label{greedy_pairing_algorithm}
    \begin{algorithmic}[1]
        \REQUIRE{distance matrix $w\in \mathbb{R}^{M\times M}$ and batch size $M$}
        \renewcommand{\algorithmicrequire}{\textbf{Output:}}
        \REQUIRE{pairing matrix $p\in {\{0,1\}}^{M\times M}$}
        \STATE Initialize $p$ as an $M\times M$ zero matrix
        \STATE $(i,j)$ = $\underset{i,j}{\text{argmax}}\quad w$
        \STATE Set $p_{i,j}$ as $1$, and $i$ as $i_{\text{first}}$
        \STATE Set $w_{k,i}$ = $0$, for all $k\in \{0,...,M-1\}$
        \FOR{$k=1$ {\bfseries to} $M-2$}
            \STATE $i$ $\leftarrow$ $j$
            \STATE $j$ = $\text{argmax}\quad w_{i,:}$
            \STATE Set $p_{i,j}$ as $1$
            \STATE Set $w_{k,i}$ = $0$, for all $k\in \{0,...,M-1\} $
        \ENDFOR
        \STATE Set $p_{j,i_{\text{first}}}$ as $1$
        
    \end{algorithmic}
\end{algorithm}

\subsection{Masking Algorithm} 
To keep the salient regions in the mixup image, we use the following algorithm to generate the mask after finding the best pairs among mini-batch. 
Since the sum-to-1 normalized saliency values $z_k$ are too small in each pixel to be used as a mask, the pixel values of mixed images may abnormally decrease when they are directly multiplied by the original image as a mask. To prevent this, we adjust the pixel-wise mixing ratio of each pixel by dividing the saliency map by the sum of the source and target saliency map, denoted as $\oslash$. With the pixel-wise mixing ratio, a mask can be assigned appropriately from both the source and target saliency map that the pixels are desired to highlight. In other words, each pixel in the source saliency map will have high expressive power in the mixed images when it is larger than the corresponding pixel in the target saliency map. Comparing saliency maps in a pixel-wise manner can lead to the salient region being highlighted, as shown in Figure \ref{method_images}. The following Mixup equation is the function of adjusting the pixel-wise mixing ratio matrix between source and target masks.
\begin{equation} \label{mixup_eq_individual}
    h(x_{s},x_{t})= {z_{s}}\oslash{(z_{s}+z_{t})}\odot x_{s} +{z_{t}}\oslash{(z_{s}+z_{t})}\odot x_{t},
\end{equation}
where $z_{s},z_{t} \in \mathbb{R}^{N}$
are sum-to-1 normalized saliency maps. Given input $x_{s}$ and $x_{t}$, our mixup function determines the final mixup masks that are generated by utilizing pixel-wise mixing ratio determination between two sum-to-1 normalized saliency maps $z_{t}$ and $z_{s}$. 
Given mixup masks, the ground-truth of a mixed image $\hat{y}$ is determined as below:
\begin{equation}
    \hat{y}_{(s,t)}= \frac{1 }{ N}\sum_{k\in N}{ \frac{z_{s,k}}{z_{s,k}+z_{t,k}}}y_{s} + \frac{1}{N}\sum_{k\in N}{ \frac{z_{t,k}}{z_{s,k}+z_{t,k}}  }y_{t}.
\end{equation}
Finally, with the pairing matrix $p$ from Section \ref{pairing_algorithm} and the input batch $x_B$ as an input of mixup function, we can reformulate Equation \ref{mixup_eq_individual} to our final mixup function as below:
\begin{equation} \label{mixup_eq_batch}
    \begin{aligned}
    h(x_{B})= & {z_{B}}\oslash{(z_{B}+p^{\top}z_{B})}\odot x_{B}\\
             & + {p^{\top}z_{B}}\oslash{(z_{B}+p^{\top}z_{B})}\odot p^{\top}x_{B}.
    \end{aligned}
\end{equation}

In summary, our framework is two-fold. First, we determine the pairs following the proposed pairing algorithm with a given batch. Second, after the blur and normalization to the saliency maps, we generate mixup images by linearly interpolating original images weighted by mixup masks, which are pixel-wise mixing ratio matrices.

\section{Experiments}
In this section, we evaluate the performance and efficiency of GuidedMixup. First, we compare the generalization performance of the proposed method against baselines by training classifiers on CIFAR-100 \cite{Krizhevsky09learningmultiple}, Tiny-ImageNet \cite{chrabaszcz2017downsampled}, and ImageNet \cite{deng2009imagenet} datasets. To validate a broader impact on the generalization, we also measure the performance of four Fine-Grained Vision Classification (FGVC) datasets, which are Caltech-UCSD Birds-200-2011 (CUB) \cite{WahCUB_200_2011}, Stanford Cars (Cars) \cite{KrauseStarkDengFei-Fei_3DRR2013}, FGVC-Aircraft (Aircraft) \cite{maji2013fine}, and Caltech-101 (Caltech) \cite{fei2006one}. To demonstrate the efficiency of our method, we measure the augmentation overhead of all the mixup strategies and examine the generalization performance and overhead simultaneously. Furthermore, we show that our proposed method can improve the performance and robustness simultaneously with AugMix \cite{hendrycks2020augmix}, which is proposed to improve robustness and uncertainty. 
\begin{table}[t]
    \centering
    \small{
    \begin{tabular}{lcccc}
    \toprule
    \multirow{3}{*}{Method} &\multicolumn{2}{c}{\textbf{CIFAR-100}} & \multicolumn{2}{c}{\textbf{Tiny-ImageNet}} \\
    \cmidrule{2-5}
                     &   Top-1        & Top-5         &  Top-1          & Top-5          \\
                     &   Err (\%)     & Err (\%)      &  Err (\%)       & Err (\%)     \\
    \midrule
    Vanilla                & 23.67          & 8.98          & 42.77           & 26.35           \\
    Input Mixup            & 23.16          & 7.58          & 43.41           & 26.98           \\
    Manifold Mixup         & 20.98          & 6.63          & 41.99           & 25.88           \\
    CutMix                 & 23.20          & 8.09          & 43.33           & 26.41           \\
    AugMix                 & 24.69          & 8.38          & 44.03           & 25.32           \\ 
    SaliencyMix $\ddagger$ & 20.25          & \textbf{5.29} & 43.46           & 23.86           \\
    Guided-SR              & \textbf{19.40} & 6.00          & \textbf{40.56}  & \textbf{23.46}  \\ 
    \midrule
    PuzzleMix              & 19.62          & 5.85          & 36.52           & 24.48          \\
    Co-Mixup $\dagger$     & 19.85          & -             & 35.85           & -                  \\
    Guided-AP              & \textbf{18.80} & \textbf{5.12} & \textbf{35.37}  & \textbf{17.51} \\ 
    \bottomrule
    \end{tabular}
    }
    \caption{Top-1 / Top-5 error rates (\%, $\downarrow$) using PreActResNet-18 on CIFAR-100, and Tiny-ImageNet. The results of Co-Mixup$\dagger$ are from \citet{kim2020co}. We reproduce the results of SaliencyMix$\ddagger$. The rest are reported on PuzzleMix \cite{kim2020puzzle}.}
    \label{generalization}
\end{table}

\subsection{Implementation Details} \label{implementation_details}

GuidedMixup can be used with any saliency detection method. In this paper, we evaluate our method with two saliency detection approaches; a spectral residual approach for efficiency \cite{hou2007saliency} and another approach \cite{simonyan2013deep}, for a fair comparison with \citet{kim2020puzzle,kim2020co}, which requires higher complexity due to additional forward and backward passes to obtain precise saliency. We denote GuidedMixup using each saliency detection method as Guided-SR and Guided-AP. After the saliency detection, we employ Gaussian blur to obtain a saliency map that covers the salient region well. We use kernel size as 7 and the sigma of Gaussian distribution as $\sigma$=3 in both our methods.

We categorize all the mixup strategies into two groups: whether requiring additional propagation of model in mixup strategies or not. One group is SnapMix, PuzzleMix, Co-Mixup, and Guided-AP, which require higher computational costs, and all the other mixup methods belong to the other group in the following experiments. Note that SnapMix utilizes class activation map \cite{zhou2016learning} as a saliency detection method, which requires only a forward pass. The other mixup methods in the group requiring additional propagation utilize \citet{simonyan2013deep}, which requires both forward and backward passes.

\subsection{Generalization Performance on General Classification Datasets}
To validate the generalization performance of our GuidedMixup, we first train and evaluate the model on classification datasets such as CIFAR-100, Tiny-ImageNet, and ImageNet. First, we evaluate our method on CIFAR-100 and Tiny-ImageNet with a residual network model (PreActResNet-18) \cite{he2016identity} following the protocol of \citet{verma2019manifold}. As shown in Table \ref{generalization}, we observe that Guided-SR outperforms the other augmentation baselines among the group, which does not require additional network propagations. Guided-AP also achieves the best performance in generalization on CIFAR-100 and Tiny-ImageNet among all the mixup strategies. ImageNet will be discussed in Section \ref{corrupted and scarce}. 

\begin{figure}[t]
    \centering
    \includegraphics[width=1\linewidth]{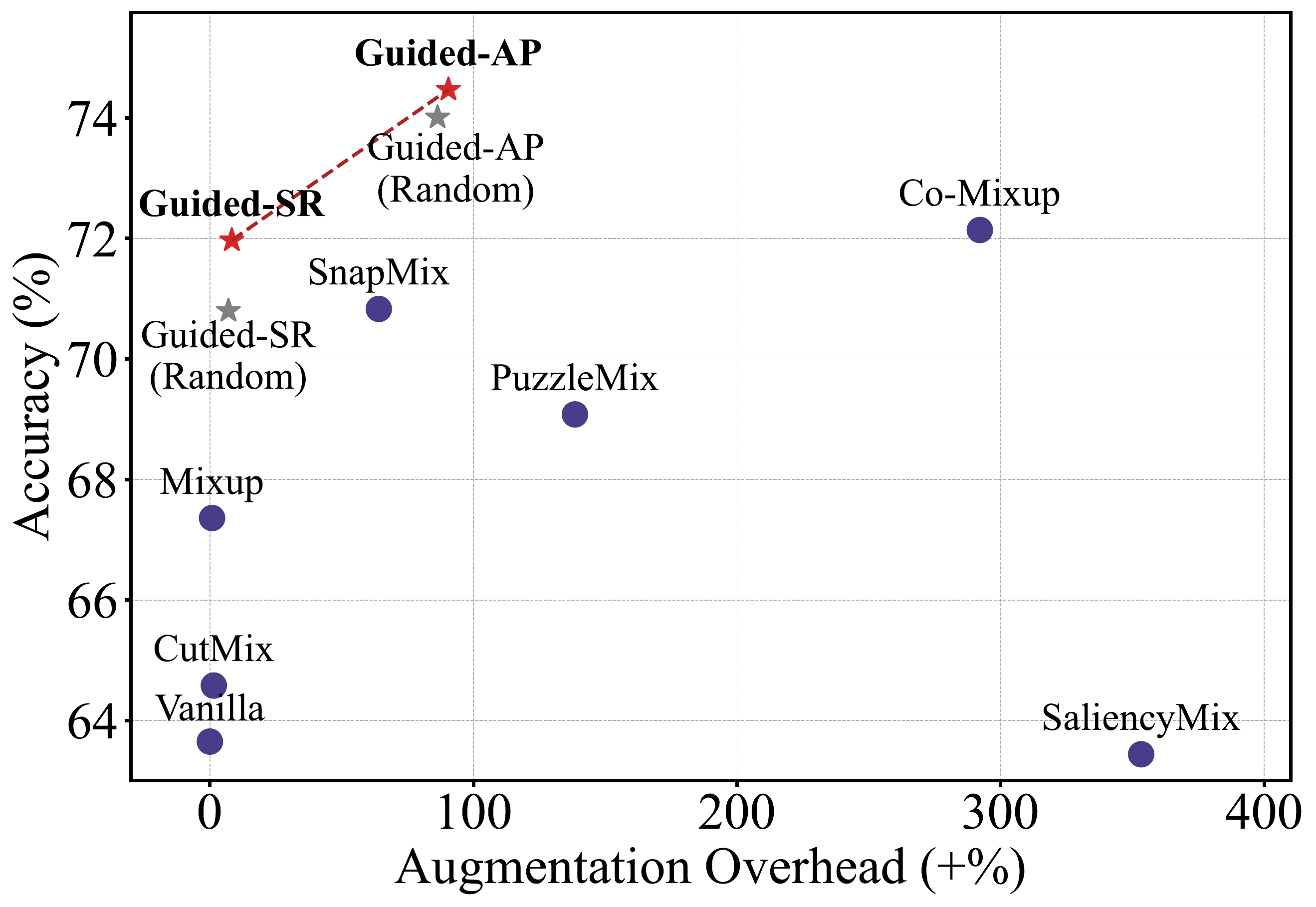}
    \caption{Augmentation overhead (+\%) - accuracy (\%) plot on CUB dataset with batch size 16. The closer to the upper left corner, the better the augmentation strategy is.} 
    \label{subfigtab:batchtimeratio_images}
\end{figure}

\begin{table}[t]
\centering
\begin{tabular}{p{0.13\columnwidth}>{\centering}p{0.09\columnwidth}>{\raggedleft}p{0.12\columnwidth}>{\raggedleft}p{0.12\columnwidth}>{\raggedleft\arraybackslash}p{0.12\columnwidth}}
    \toprule
    \multirow{2}{*}{Method} &  \multicolumn{1}{|c|}{Batch size} &  16  &  32  &  64 \\
    \cmidrule{2-5}
                         &  &\multicolumn{3}{|c}{Augmentation overhead (+\%)} \\
    \midrule
    \multicolumn{2}{l}{ Mixup}              & 0.9   & 0.6 & 0.4          \\
    \multicolumn{2}{l}{ CutMix}             & 1.5   & 1.0 & 0.6          \\
    \multicolumn{2}{l}{ SaliencyMix}        & 353.3 & 701.8 & 923.3          \\
    \multicolumn{2}{l}{ Guided-SR (Random)} & 6.4   & 4.2 & 2.7          \\
    \multicolumn{2}{l}{ Guided-SR (Greedy)} & 7.7   & 7.1 & \textbf{7.0} \\ 
    \midrule
    \multicolumn{2}{l}{ SnapMix}            & 67.4  & 64.9 & 60.2          \\ 
    \multicolumn{2}{l}{ PuzzleMix}          & 138.5  & 139.9 & 134.1          \\ 
    \multicolumn{2}{l}{ Co-Mixup}           & 292.1  & 490.2 & 716.6          \\
    \multicolumn{2}{l}{ Guided-AP (Random)} & 87.8  & 81.9 & 70.1          \\
    \multicolumn{2}{l}{ Guided-AP (Greedy)} & 89.2  & 83.0 & \textbf{77.5} \\
    \bottomrule
\end{tabular}
\captionof{table}{Augmentation overhead ($\downarrow$, +\%) of mixup methods with three different mini-batch sizes on CUB dataset.} 
\label{subfigtab:speed_per_batch}
\end{table}

\subsection{Augmentation Overhead}
\label{section:augmentation-overhead}
In this section, we compare the computational efficiency of mixup strategies in addition to their generalization performance. To evaluate the computational efficiency, we utilize the augmentation overhead, which is the ratio of training time increased by including the augmentation as follows:
$$Augmentation\text{ }overhead=\frac{\textit{T}_{aug}-\textit{T}_{vanilla}}{\textit{T}_{vanilla}}\times 100 (\%),$$
where \textit{T} is the total training time, and $\textit{T}_{vanilla}$ is the training time without augmentation.
In this evaluation, we compare the augmentation overhead without multi-processing during the augmentation process for a fair comparison.

\begin{table*}[t!]
    \centering
    \small{
    \begin{tabular}{lcccccccccccc}
    \toprule
    \multirow{2}{*}{Method} & \multicolumn{4}{|c}{ResNet-18} & \multicolumn{4}{|c}{ResNet-50} & \multicolumn{4}{|c}{DenseNet-121}   \\
    \cmidrule{2-13}
    & \multicolumn{1}{|c}{CUB} & Cars & Aircraft & Caltech & \multicolumn{1}{|c}{CUB}& Cars & Aircraft & Caltech & \multicolumn{1}{|c}{CUB}& Cars & Aircraft & Caltech\\
    \midrule
    Vanilla           & 63.65          & 81.85          & 77.28          & 88.42          & 65.50          & 85.52          & 80.29          & 87.50         & 69.35          & 85.45          & 79.96          & 89.29          \\
    Mixup             & 67.36          & 85.14          & 79.39          & 90.15          & 71.33          & 88.14          & 82.38          & 91.01         & 74.23          & 89.06          & 81.73          & 91.47          \\
    CutMix            & 63.64          & 86.06          & 79.90          & 89.52          & 72.58          & 89.22          & 82.45          & 91.24         & 74.30          & 88.84          & 81.43          & 91.36          \\
    SaliencyMix       & 63.48          & 86.32          & 80.05          & 89.63          & 66.66          & 89.04          & 83.14          & 91.30         & 68.75          & 88.91          & 80.86          & 90.78          \\
    \textbf{Guided-SR}& \textbf{71.95} & \textbf{86.50} & \textbf{80.94} & \textbf{90.66} & \textbf{74.08} & \textbf{89.23}         & \textbf{83.51} & \textbf{91.36} & \textbf{76.51} & \textbf{89.40} & \textbf{83.22} & \textbf{91.82} \\
    \midrule
    SnapMix           & 70.83          & 87.45          & 80.64          & 89.75          & 75.53          & 90.10          & 82.96          & 90.78         & 75.76          & 90.27          & 83.29          & 91.36\\
    PuzzleMix         & 69.08          & 87.27          & 79.18          & 90.44          & 74.85          & 89.68          & 82.66          & 90.53         & 77.27          & 90.10          & 82.63          & 91.47 \\ 
    Co-Mixup          & 72.14          & 87.59          & 80.23          & 90.03          & 72.83          & 89.53          & 83.57          & 90.38         & 77.05          & 90.23          & 83.35          & 90.44 \\
    \textbf{Guided-AP}& \textbf{74.47} & \textbf{87.71} & \textbf{81.58} & \textbf{91.32} & \textbf{77.08} & \textbf{90.27} & \textbf{84.32} & \textbf{91.49}& \textbf{77.52} & \textbf{90.47} & \textbf{84.09} & \textbf{91.94} \\ 
    \bottomrule
    \end{tabular}
    }
    \caption{Top-1 accuracy rate (\%, $\uparrow$) on FGVC datasets using ResNet-18, ResNet-50, and DenseNet-121.}
    \label{fgvc_acc}
\end{table*}

Figure \ref{subfigtab:batchtimeratio_images} clearly shows the efficiency of our methods and confirms that both Guided-SR and Guided-AP are good trade-off solutions between generalization performance and augmentation overhead. 
Both methods enhance the accuracy rate up to 2\% in each group, and Guided-SR is better than some methods with much bigger overhead. Also, we compare the augmentation overhead with three mini-batch sizes (16, 32, and 64) in Table \ref{subfigtab:speed_per_batch}.
Another thing to note is that the overhead of Co-Mixup is about four times higher than that of Guided-AP, which utilizes the same saliency detection method, even with a small batch size of 16. 

\subsection{Generalization Performance on Fine-Grained Vision Classification (FGVC) Datasets}
Classifying Fine-Grained Vision Classification (FGVC) datasets is a challenging task since the model needs to discriminate fine differences in the object. To check whether our method preserves the subtle object parts well, we compare GuidedMixup with the mixup baselines on four standard FGVC datasets using ResNet \cite{he2016deep} and DenseNet \cite{huang2017densely} architectures. Since previous mixup methods do not provide official reports on FGVC datasets, we reproduce their results based on the released codes. As shown in Table \ref{fgvc_acc}, our proposed methods improve the generalization performance with a large margin in each group. Especially, Guided-AP achieves \emph{state-of-the-art} performance on all datasets.

Comparing the results of Guided-SR and Guided-AP in Table \ref{generalization} and Table \ref{fgvc_acc}, Guided-AP consistently outperforms Guided-SR. This reconfirms that saliency detection in Guided-AP provides more precise saliency information than the spectral residual approach because it delivers information on what parts the model really cares about.

\begin{table}[t]
    \centering
    \begin{tabular}{lccc}
        \toprule
        Method & Top-1 Err (\%)         & Corruption Err (\%) \\
        \midrule 
        Vanilla              & 21.14          & 49.08          \\ 
        AugMix               & 20.45          & 32.22          \\
        Guided-SR            & 16.54          & 42.89          \\
        Guided-SR (Aug)      & \textbf{15.96} & \textbf{29.54} \\
        \midrule
        PuzzleMix            & \textbf{15.95} & 42.46          \\
        PuzzleMix (Aug)      & 16.60          & 29.91          \\
        Guided-AP            & 16.37          & 41.11          \\
        Guided-AP (Aug)      & 15.98          & \textbf{29.07} \\
    \bottomrule
    \end{tabular}
\caption{Top-1 / mean Corruption Error rates (\%, $\downarrow$) using WideResNet28-10 on CIFAR-100 and CIFAR-100-C. All the results are from PuzzleMix \cite{kim2020puzzle}.}
\label{robustness}
\end{table}

\begin{table}[t]
    \small{
    \begin{tabular}{lcc|cc}
        \toprule
        \multirow{3}{*}{Method} & \multicolumn{2}{c|}{Clean} & \multicolumn{2}{c}{Corruption Type}        \\
        \cmidrule{2-5}
                                & Top-1    & Top5     & Gaussian & Random      \\
                                & Err (\%) & Err (\%) & Noise    & Replacement \\
        \midrule
        Vanilla         & 24.03          & 7.34          & 29.12          & 41.73          \\
        Input           & 22.97          & 6.48          & \textbf{26.29} & 39.41          \\
        CutMix          & 22.92          & 6.55          & 27.11          & 46.20          \\
        AugMix$\dagger$ & 23.25          & 6.70          & -              & -              \\
        Guided-SR       & \textbf{22.80} & \textbf{6.34} & 26.52          & \textbf{38.84} \\
        \midrule
        PuzzleMix       & 22.49          & 6.24          & 26.11          & 39.23          \\
        Co-Mixup        & \textbf{22.37} & 6.16          & 25.89          & 38.77          \\
        Guided-AP       & 22.47          & \textbf{6.14} & \textbf{24.66} & \textbf{38.72} \\
        \bottomrule
    \end{tabular} 
    }
    \caption{Top-1 and Top-5 error rates (\%, $\downarrow$) on clean ImageNet dataset and Top-1 error rate (\%) on corrupted ImageNet validation set \cite{kim2020co} using ResNet-50. All the results are from PuzzleMix \cite{kim2020puzzle}, and Co-Mixup \cite{kim2020co}. The results of AugMix$\dagger$ on the corrupted dataset are not reported in the original paper.}
    \label{imagenet_corrupted}
\end{table}
\subsection{Robustness on Corrupted Dataset and Scarce Dataset} \label{corrupted and scarce}
\textbf{Data Corruption} Machine learning techniques assume that the training data distribution is similar to the test data distribution. Therefore, small corruption on the test data might cause mis-inference \cite{lee2020network}. In response to this issue, AugMix \cite{hendrycks2020augmix} is one of the augmentation methods which cope well with corrupted images while improving generalization performance.
In this section, we evaluate the robustness of our methods on corrupted datasets. Motivated by \citet{kim2020puzzle}, we also tested GuidedMixup in combination with AugMix, which is an augmentation method that copes with corrupted images well. We simply use AugMix images as input for our methods, denoted as Guided-SR (Aug) and Guided-AP (Aug). CIFAR-100-C consists of 19 different corruptions (including blur, noise, weather, and digital corruption types), and we measure the average test error. In Table \ref{robustness}, we find that PuzzleMix with AugMix sacrifices the generalization performance to improve robustness. However, Guided-SR and Guided-AP combined with AugMix improve both generalization performance and robustness.

Next, we train ResNet-50 on ImageNet dataset for 100 epochs following the protocol of \citet{kim2020co} for evaluating both generalization and robustness. As shown in the `clean' column of Table \ref{imagenet_corrupted}, Guided-AP is the second best, next to Co-Mixup, which has a significantly larger augmentation overhead as shown in Table \ref{section:augmentation-overhead}. Considering this big difference in augmentation overhead, we believe Guided-AP provides a meaningful trade-off point even on ImageNet.

Also, there is the corrupted ImageNet validation set from \citet{kim2020co}, which consists of the following two types of corrupted images; images partially random-replaced by other images and images distorted by random Gaussian noise. We evaluate the test errors on each corrupted dataset using ResNet-50 trained with each mixup strategy. As shown in Table \ref{imagenet_corrupted}, Guided-AP outperforms other mixup baselines for each corrupted dataset.

\textbf{Data Scarcity} Deep neural networks have suffered from data scarcity, which often causes overfitting. Because the primary purpose of augmentation is to regularize the model by effectively increasing the number of data, we evaluate the generalization performance under insufficient dataset conditions.
We compare the accuracy of WideResNet28-10 \cite{zagoruyko2016wide} trained with mixup strategies using only 10\%, 20\%, and 50\% of the original CIFAR-100 dataset. In Table \ref{data_scarcity}, both Guided-SR and Guided-AP show superior effectiveness when training data is more scarce. Moreover, we evaluate all the mixup baselines on Oxford 102 Flower dataset \cite{nilsback2008automated}, which has only 10 training images per class. Table \ref{flowers} shows that our GuidedMixup consistently outperforms other mixup methods in situations where data is limited.


\begin{table}[t!]
    \centering
    \begin{tabular}{lccc}
        \toprule
        \multirow{2}{*}{Method} & \multicolumn{3}{c}{The number of data per class} \\
                                & 50 (10\%) & 100 (20\%) & 250 (50\%)     \\
        \midrule 
        Vanilla   & 40.10 \small{(0.70)}          & 55.56 \small{(0.21)}          & 70.16 \small{(0.17)}           \\ 
        Input     & 49.44 \small{(0.65)}          & 61.74 \small{(1.67)}          & 74.31 \small{(0.26)}          \\
        Manifold  & 47.94 \small{(0.70)}          & 62.25 \small{(0.43)}          & 75.00 \small{(0.16)}          \\
        CutMix    & 42.81 \small{(1.40)}          & 60.14 \small{(0.81)}          & 74.94 \small{(0.70)}          \\
        Guided-SR & \textbf{50.60} \small{(0.48)} & \textbf{63.99} \small{(0.49)} & \textbf{75.57} \small{(0.10)} \\
        \midrule
        PuzzleMix & 50.13 \small{(0.84)}          & 63.99 \small{(0.16)}          & 76.31 \small{(0.27)}          \\
        Co-Mixup  & 50.50 \small{(0.98)}          & 64.47 \small{(0.63)}          & 75.43 \small{(0.12)}          \\
        Guided-AP & \textbf{54.25} \small{(0.49)} & \textbf{66.22} \small{(0.50)} & \textbf{76.70} \small{(0.10)} \\
    \bottomrule
    \end{tabular}
    \caption{Top-1 accuracy rates (\%, $\uparrow$) on CIFAR-100 with a reduced number of data per class using WideResNet28-10. The values in parentheses are the standard deviation of Top-1 accuracy rates. We experiment with three different seeds.}
    \label{data_scarcity}
\end{table}

\begin{table}[t!]
    \centering
    \begin{tabular}{lccc}
        \toprule
        Method & Validation Acc (\%) & Test Set Acc (\%)  \\
        \midrule 
        Vanilla           & 64.48 (1.60)          & 59.14 (1.15)   \\ 
        Mixup             & 70.55 (0.91)          & 66.81 (0.43)   \\
        CutMix            & 62.68 (2.20)          & 58.51 (0.42)   \\
        SaliencyMix       & 63.23 (1.19)          & 57.45 (1.00)   \\
        Guided-SR         & \textbf{72.84 (0.52)} & \textbf{69.31 (0.23)} \\
        \midrule
        SnapMix           &        65.71 (1.56)   & 59.79 (0.64)   \\
        PuzzleMix         &        71.56 (0.79)   & 66.71 (0.10)   \\
        Co-Mixup          &        68.17 (0.54)   & 63.20 (0.36)   \\
        Guided-AP         & \textbf{74.74 (0.40)} & \textbf{70.44 (0.69)}   \\
    \bottomrule
    \end{tabular}
\caption{Top-1 accuracy rates (\%, $\uparrow$) on Flower dataset using ResNet-18. The values in parentheses are the standard deviation of Top-1 accuracy rates. We experiment with three different seeds.} 
\label{flowers}
\end{table}

\subsection{Impact of Pairing Algorithm} \label{impact_pairing}
GuidedMixup utilizes the proposed pairing algorithm in all the experiments to find good pairs among input images by utilizing a saliency map without handling a mixup mask. Therefore, This algorithm can also be applied to other mixup-based augmentation strategies. To verify the effectiveness of our pairing algorithm, we compare the performance of mixup baselines to whether the proposed pairing algorithm exists or not (in Table \ref{difference_pairing}). The results indicate that our pairing algorithm is effective not only for our GuidedMixup, but also for most other mixup-based methods. 
\begin{table}[t]
    \centering
        \small{
        \begin{tabular}{lccc}
            \toprule
            \multirow{2}{*}{Method} & \multicolumn{3}{c}{Top-1 Accuracy (\%)}\\
            \cmidrule{2-4}
            & TinyImageNet & CUB & Aircraft \\
            \midrule
            CutMix      & 57.81(+1.14) & 64.41(+0.77) & 80.09(+0.19) \\
            SaliencyMix & 58.19(+1.65) & 63.88(+0.40) & 80.37(+0.31) \\
            Guided-SR   & 59.44(+0.51) & 71.95(+1.15) & 80.94(+0.44) \\
            \midrule
            PuzzleMix   & 63.82(+0.34) & 70.47(+1.49) & 79.89(+0.71) \\ 
            Guided-AP   & 64.63(+0.26) & 74.47(+0.78) & 81.58(+0.19) \\ 
            \bottomrule
        \end{tabular}
        }
    \caption{Top-1 accuracy rate (\%, $\uparrow$) of the mixup methods with the proposed pairing algorithm on CUB and Aircraft datasets using ResNet-18 and on Tiny-ImageNet using PreActResNet-18. The values in parentheses are the improved accuracy by using our pairing algorithm.}
    \label{difference_pairing}
\end{table}

\section{Conclusion}
We presented GuidedMixup, which generates a mixup image with objects from input images well preserved by efficiently determining mixup pairs and masks using the saliency information. Given mini-batch images, GuidedMixup selects pairs with minimally overlapping salient regions by an efficient and effective pairing algorithm. When mixing the chosen images, GuidedMixup adjusts the pixel-wise mixing ratio to maintain the salient regions of each image. 
Compared to previous mixup-based augmentation methods, GuidedMixup achieves the state-of-the-art generalization performance on CIFAR-100, Tiny-ImageNet, and all Fine-Grained Visual Classification datasets we tested, with \emph{much less computational overhead} than previous optimization-based methods. 
In the data scarcity test, GuidedMixup performed better when the available training data gets smaller, reinforcing the merit of our method when data augmentation is more needed. 
Furthermore, our pairing algorithm can be independently used with other mixup strategies, leading to noticeable accuracy improvement. This also supports that maintaining salient regions is important in augmentation methods. 

\section*{Acknowledgements}
This work was supported by the National Research Council of Science \& Technology (NST) grant by the Korea government (MSIT) [CRC-20-02-KIST], and by Institute of Information \& Communications Technology Planning \& Evaluation (IITP) grant funded by the Korea government (MSIT) (No.2021-0-00456, Development of Ultra-high Speech Quality Technology for Remote Multi-speaker Conference System).
\bibliography{aaai23.bib}

\end{document}